\documentclass[runningheads]{llncs}
\usepackage{graphicx}
\usepackage{todonotes}
\usepackage{url}
\usepackage{cite}
\usepackage{bbm}
\usepackage{amsmath}
\usepackage{amssymb}
\usepackage{multirow}
\usepackage{algorithm,algorithmic}
\usepackage{comment}
\usepackage{graphicx}
\usepackage{bbm}
\usepackage{bm}
\usepackage{dsfont}
\usepackage{mathtools}
\usepackage{adjustbox}
\usepackage{hyperref}

\begin{document}
\title{Compositional Representation Learning for Brain Tumour Segmentation}
\titlerunning{Compositional Representation Learning for Brain Tumor Segmentation}

\author{
Xiao Liu\inst{1,2}, Antanas Kascenas\inst{1}, Hannah Watson\inst{1}, Sotirios A. Tsaftaris\inst{1,2,3} \and Alison Q. O'Neil\inst{1,2}}
%index{Liu, Xiao}
%index{Kascenas, Antanas}
%index{Watson, Hannah}
%index{Tsaftaris, Sotirios A.}
%index{O'Neil, Alison Q.}
\authorrunning{X. Liu et al.}
\institute{Canon Medical Research Europe Ltd., Edinburgh, UK 
\and
School of Engineering, University of Edinburgh, Edinburgh EH9 3FB, UK 
\and
% University of Glasgow, Glasgow G12 8QQ, UK
% \and
The Alan Turing Institute, London, UK \\
\email{xiao.liu@mre.medical.canon}}

\maketitle              % typeset the header of the contribution
\sloppy
\begin{abstract}
For brain tumour segmentation, deep learning models can achieve human expert-level performance given a large amount of data and pixel-level annotations. However, the expensive exercise of obtaining pixel-level annotations for large amounts of data is not always feasible, and performance is often heavily reduced in a low-annotated data regime. To tackle this challenge, we adapt a mixed supervision framework, vMFNet, to learn robust compositional representations using unsupervised learning and weak supervision alongside non-exhaustive pixel-level pathology labels. In particular, we use the BraTS dataset to simulate a collection of 2-point expert pathology annotations indicating the top and bottom slice of the tumour (or tumour sub-regions: peritumoural edema, GD-enhancing tumour, and the necrotic / non-enhancing tumour) in each MRI volume, from which weak image-level labels that indicate the presence or absence of the tumour (or the tumour sub-regions) in the image are constructed. Then, vMFNet models the encoded image features with von-Mises-Fisher (vMF) distributions, via learnable and compositional vMF kernels which capture information about structures in the images. We show that good tumour segmentation performance can be achieved with a large amount of weakly labelled data but only a small amount of fully-annotated data. Interestingly, emergent learning of anatomical structures occurs in the compositional representation even given only supervision relating to pathology (tumour).

\keywords{Compositionality \and Representation learning \and Semi-supervised \and Weakly-supervised \and Brain tumour segmentation.}
\end{abstract}

\section{Introduction}
When a large amount of labelled training data is available, deep learning techniques have demonstrated remarkable accuracy in medical image segmentation \cite{antonelli2022medical}. However, performance drops significantly when insufficient pixel-level annotations are available \cite{liu2020disentangled, liu2021semi, thermos2021controllable}. By contrast, radiologists learn clinically relevant visual features from ``weak'' image-level supervision of seeing many medical scans \cite{alexander2020radiologists}.
%radiologists learn clinically relevant visual features by viewing thousands of medical images. Yet the precise visual features that expert radiologists use in their clinical practice remain unknown. 
%can quickly learn to locate the anatomy or lesion of interest from only a few examples.
% For instance, clinical experts are trained to remember relevant configurations (components) of anatomical and pathological structures from medical images they have seen before.
When searching for anatomy or lesions of interest in new images, they look for characteristic configurations of these clinically relevant features (or components). A similar compositional learning process has been shown to improve deep learning model performance in many computer vision tasks \cite{tokmakov2019learning, huynh2020compositional, kortylewski2020compositional} but has received limited attention in medical applications.

In this paper, we consider a limited annotation data regime where few pixel-level annotations are available for the task of brain tumour segmentation in brain MRI scans. Alongside this, we construct slice-level labels for each MRI volume indicating the presence or absence of the tumour.
%for the task of whole tumour segmentation and the presence or absence of the tumour sub-regions for the task of tumour sub-region segmentation.
These labels can be constructed from 2-point expert pathology annotations indicating
the top and bottom slices of the tumour, which are fast to collect. We consider that pathology annotations are not only better suited to the task (tumour segmentation) but also to the domain (brain MRI) than the originally proposed weak supervision with anatomy annotations \cite{liu2023compositionally}; annotating the top and bottom slices for anatomical brain structures such as white matter, grey matter and cerebrospinal fluid (CSF) would be relatively uninformative about the configurations of structures within the image due to their whole brain distributions.
% Note that our approach can be extended to a similar setting considered in Sli2Vol \cite{yeung2021sli2vol}, where one slice has a full annotation and other slices of the same volume can be weakly labelled.

For the learning paradigm, we investigate the utility of learning compositional representations in increasing the annotation efficiency of segmentation model training. Compositional frameworks encourage identification of the visible semantic components (e.g. anatomical structures) in an image, requiring less explicit supervision (labels). We follow \cite{kortylewski2020compositional, liu2022vmfnet, liu2023compositionally} in modelling compositional representations of medical imaging structures with learnable von-Mises-Fisher (vMF) kernels. The vMF kernels are learned as the cluster centres of the feature vectors of the training images, and the vMF activations determine which kernel is activated at each position. On visualising kernel activations, it can be seen that they approximately correspond to human-recognisable structures in the image, lending interpretability to the model predictions.
%With such weak labels, the compositional representations are constrained to better learn corresponding anatomical and pathological structures with more data that are not fully labelled. 
Our contributions are summarised as:
\begin{itemize}
    \item We refine an existing mixed supervision compositional representation learning framework, vMFNet, for the task of brain tumour segmentation, changing the weak supervision task from anatomy presence/absence to more domain-suited pathology presence/absence and simplifying the architecture and training parameters according to the principle of parsimony (in particular reducing the number of compositional vMF kernels and removing an original training subtask of image reconstruction). 
    % \item We learn better compositional kernels by introducing the image-level weak supervision task, which improves the model performance when the pixel-level data annotations are limited.
    \item We perform extensive experiments on the BraTS 2021 challenge dataset \cite{brats1, brats2, brats3} with different percentages of labelled data, showing superior performance of the proposed method compared to several strong baselines, both quantitatively (better segmentation performance) and qualitatively (better compositional representations).
    \item We compare weak pathology supervision with \emph{tumour} labels to richer tumour \emph{sub-region} labels, showing that the latter increases model accuracy for the task of tumour sub-region segmentation but also reduces the generality of the compositional representation, which loses anatomical detail and increases in pathology detail, becoming more focused on the supervision task.
\end{itemize}

\section{Related work}
Compositionality is a fundamental concept in computer vision, where it refers to the ability to recognise complex objects or scenes by detecting and combining simpler components or features \cite{lake2015human}. Leveraging this idea, compositional representation learning is an area of active research in computer vision \cite{yuille2021deep}. Early approaches to compositional representation learning in computer vision include the bag-of-visual-words model \cite{kortylewski2020combining} and part-based models \cite{kortylewski2020compositional}. Compositional representation learning has been applied to fine-grained recognition tasks in computer vision, such as recognising bird species \cite{singh2019finegan, huynh2020compositional}. In addition, compositionality has been incorporated for robust image classification \cite{tokmakov2019learning, kortylewski2020compositional} and recently for compositional image synthesis \cite{liu2021learning, arad2021compositional}. Among these works, Compositional Networks \cite{kortylewski2020compositional}, originally designed for robust classification under object occlusion, are easier to extend to pixel-level tasks as they estimate spatial and interpretable vMF likelihoods. Previous work integrates vMF kernels \cite{kortylewski2020compositional} for object localisation \cite{yuan2021robust} and recently for nuclei segmentation (with the bounding box as supervision) in a weakly supervised manner \cite{zhang2021light}. More recently, vMFNet \cite{liu2022vmfnet} applies vMF kernels for cardiac image segmentation in the domain generalisation setting. Additionally, vMFNet integrated weak labels indicating the presence or absence of cardiac structures and this gave improved performance \cite{liu2023compositionally}. We use similar types of weak image-level annotations but apply the vMF kernels to pathology segmentation and supervise with weak labels indicating the presence or absence of pathological structures. 

\begin{figure}[t]
\includegraphics[width=0.9\textwidth]{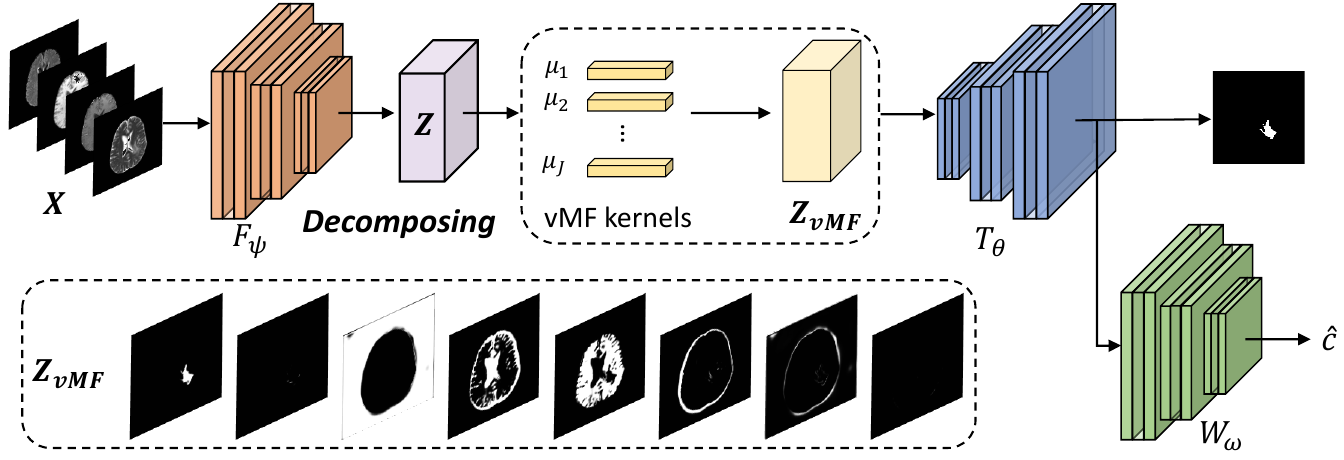}
\centering
\caption{Illustration of the brain tumour segmentation task using vMFBrain for compositional representation learning. We extract the weak supervision pathology labels (\emph{presence or absence of tumour}) from 2-point brain tumour annotations; interestingly, learning of anatomical structures somewhat emerges even without supervision. Notation is specified in Section~\ref{sec:method}.} \label{fig::model}
\end{figure}

\section{Method}
\label{sec:method}
We apply vMFNet \cite{liu2022vmfnet, liu2023compositionally}, as shown in Fig.~\ref{fig::model}, a model consisting of three modules: the feature extractor $\bm{F}_\psi$, the task network $\bm{T}_\theta$ (for brain tumour segmentation in our case), and the weak supervision network $\bm{W}_\omega$, where $\psi$, $\theta$ and $\omega$ denote the network parameters. Compositional components are learned as vMF kernels by decomposing the features extracted by $\bm{F}_\psi$. Then, the vMF likelihoods that contain spatial information are used to predict the tumour segmentation mask with $\bm{T}_\theta$. The voxel-wise output of $\bm{T}_\theta$ is also input to the weak supervision network $\bm{W}_\omega$ to predict the presence or absence of the tumour. This framework is
%shown in Fig.~\ref{fig::decomp} and
detailed below. We term our implementation \emph{vMFBrain}.

\subsection{Background: learning compositional components}
To learn compositional components, the image features $\mathbf{Z} \in \mathbb{R}^{H\times W\times D}$ are first extracted by $\bm{F}_\psi$. $H$ and $W$ are the spatial dimensions and $D$ is the number of channels. The feature vector $\mathbf{z}_i\in \mathbb{R}^{D}$ is defined as the normalised vector (i.e.~$||\mathbf{z}_i||=1$) across channels at position $i$ on the 2D lattice of the feature map. Then, the image features are modelled with $J$ vMF distributions. Each distribution has a learnable mean that is defined as vMF kernel $\boldsymbol \mu_{j} \in \mathbb{R}^{D}$. To ensure computational tractability, a fixed variance $\sigma$ is set for all distributions. The vMF likelihood for the $j^{th}$ distribution at each position $i$ is calculated as: 
\begin{equation}
    p(\mathbf{z}_i|\boldsymbol \mu_{j}) = \frac{e^{\sigma_{j} \boldsymbol \mu_{j}^{T} \mathbf{z}_i}}{C} , \text{ s.t. } ||\boldsymbol \mu_{j}||=1, 
\label{eq::vmf}
\end{equation}
where $C$ is a constant. This gives the vMF likelihood vector $\mathbf{z}_{i,vMF}\in \mathbb{R}^{J}$, a component of
%After modeling the image features with $J$ vMF distributions with Eq.\ \ref{eq::vmf},the vMF likelihoods
$\mathbf{Z}_{\textit{vMF}} \in \mathbb{R}^{H\times W\times J}$, which determines which kernel is activated at each position. To update the kernels during training, the clustering loss $\mathcal{L}_{clu}$ is defined in \cite{kortylewski2020compositional} as:
\begin{equation}
\mathcal{L}_{clu}(\boldsymbol \mu, \mathbf{Z}) = -{(HW)}^{-1}\sum_{i} \operatorname*{max}_{j} \boldsymbol \mu_{j}^T\mathbf{z}_i, \end{equation}
%where hard assignment of the feature vectors $\mathbf{z}_i$ to the vMF kernels $\boldsymbol \mu_{j}$ will be achieved during training. Here,
where the kernel $\boldsymbol \mu_{j}$ which is maximally activated for each feature vector $\mathbf{z}_i$ is found, and the distance between the feature vectors and their corresponding kernels is minimised by updating the kernels.
Overall, feature vectors in different images corresponding to the same anatomical or pathological structure will be clustered and activate the same kernels.
%In other words, the vMF kernels are learnt as the components or patterns of the anatomical or pathological parts.
Hence, the vMF likelihoods $\mathbf{Z}_{\textit{vMF}}$ for the same anatomical or pathological features in different images will be aligned to follow the same distributions (with the same means).

\subsection{vMFBrain for brain tumour segmentation}
Taking the vMF likelihoods as input, a follow-on segmentation task module $\bm{T}_\theta$, is trained to predict the tumour segmentation mask, i.e. $\mathbf{\hat Y}=\bm{T}_\theta(\mathbf{Z}_{\textit{vMF}})$.
Firstly, we use direct strong supervision from the available pixel-level annotations $\mathbf{Y}$. %Specifically, the segmentation mask tells what anatomical or pathological part the feature vector $\mathbf{z}_i$ corresponds to, which provides further guidance for the model to learn the vMF kernels as the components of the anatomical parts.
Secondly, we define the weak supervision label $c$ as a scalar (or a vector $\mathbf{c}$) which indicates the presence or absence of the tumour (or the presence or absence of the tumour sub-regions) in the 2D image slice. We use the output of the segmentation module as the input for a weak supervision classifier $\bm{W}_\omega$ i.e. $\hat{c} = \bm{W}_\omega(\mathbf{\hat Y})$.
We train the classifier using $L$1 distance i.e. $\mathcal{L}_{weak}(\hat{c}, c) = |\hat{c} - c|_1$.

Overall, the model contains trainable parameters $\psi$, $\theta$, $\omega$ and the vMF kernel means $\boldsymbol \mu$. The model (including all the modules) is trained \textbf{end-to-end} with the following objective:
\begin{equation}
    \operatorname*{argmin}_{\psi, \theta, \omega, \boldsymbol \mu}
    \mathcal{L}_{clu} +
    \lambda_{Dice} \mathcal{L}_{Dice}(\mathbf{Y}, \hat{\mathbf{Y}})(\boldsymbol \mu, \mathbf{Z}) + \lambda_{weak}\mathcal{L}_{weak} (\hat{c}, c),
    \label{Eqa::totalloss}
\end{equation}
where $\mathcal{L}_{Dice}$ is Dice loss \cite{milletari2016v, dice1945measures}. We set $\lambda_{Dice}=1$ when the ground-truth mask $\mathbf{Y}$ is available, otherwise $\lambda_{Dice}=0$. We set $\lambda_{weak}$ as 0.5 for the whole tumour segmentation task and $\lambda_{weak}$ as 0.1 for the tumour sub-region segmentation task (values determined empirically).

\section{Experiments}

\subsection{Dataset}
We evaluate on the task of brain tumour segmentation using data from the BraTS 2021 challenge \cite{brats1, brats2, brats3}. This data comprises native (T1), post-contrast T1-weighted (T1Gd), T2-weighted (T2), and T2 Fluid Attenuated Inversion Recovery (FLAIR) modality volumes for 1,251 patients from a variety of institutions and scanners. We
%follow \cite{kascenas2023role} in splitting
split the data into train, validation and test sets containing 938, 62 and 251 subjects.
The data has already been co-registered, skull-stripped and interpolated to the same resolution, each volume having 155 2D slices. Labels are provided for tumour sub-regions: the peritumoural edema (ED), the GD-enhancing tumour (ET), and the necrotic and non-enhancing tumour (NE). We additionally downscale all images to $128\times 128$. 

\subsection{Baselines} We compare to the baselines \textbf{UNet} \cite{ronneberger2015u}, \textbf{SDNet} \cite{chartsias2019disentangled} and \textbf{vMFNet} \cite{liu2022vmfnet}.
%It contains skip connections between the encoder and decoder layers.
\textbf{SDNet \cite{chartsias2019disentangled}} is a semi-supervised disentanglement model with anatomy and modality encoders to separately encode the anatomical structure information and the imaging characteristics. The anatomical features are used as the input to the segmentor for the task of segmentation; the model is also trained with unlabelled data on the task of reconstructing the image by recombining the anatomy and modality factors. We compare to \textbf{vMFNet} with the architecture and training loss as described in \cite{liu2022vmfnet}; this setup does not use weak supervision and has an additional image reconstruction module which we found empirically not to help performance (which thus we omit from vMFBrain). 

%Apart from comparing the model performance quantitatively, we qualitatively compare the interpretability and compositionality of representations learnt by SDNet, vMFNet and vMFBrain.

\subsection{Implementation}

\noindent \textbf{Imaging backbone:} $\bm{F}_\psi$ is a 2D UNet \cite{ronneberger2015u} (without the output classification layer) to extract features $\mathbf{Z}$. The four modalities are concatenated as the input (with 4 channels) to $\bm{F}_\psi$. For a fair comparison, we use this same UNet implementation as the backbone for all models.

\vspace{2pt} \noindent \textbf{vMFNet and vMFBrain\footnote{The code for vMFNet is available at \url{https://github.com/vios-s/vMFNet}.}:} $\bm{T}_\theta$ is a shallow convolutional network. $\bm{W}_\omega$ is a classifier model. Following \cite{kortylewski2020compositional}, we set the variance of the vMF distributions as 30. The number of kernels is set to 8, as this number performed best empirically in our early experiments. For vMF kernel initialisation, we pre-train a 2D UNet for 10 epochs to reconstruct the input image with all the training data. After training, we extract the corresponding feature vectors and perform k-means clustering, then use the discovered cluster centres to initialise the vMF kernels.

\vspace{2pt} \noindent \textbf{Training:} All models are implemented in PyTorch \cite{paszke2019pytorch} and are trained using an NVIDIA 3090 GPU. Models are trained using the Adam optimiser \cite{kingma2014adam} with a learning rate of $1\times e^{-4}$ using batch size 32. In semi-supervised and weakly supervised settings, we consider the use of different percentages of fully labelled data to train the models. For this purpose, we randomly sample 2D image slices and the corresponding pixel-level labels from the whole training dataset.

\begin{table}[t]
\centering
\caption{Dice (\%) and Hausdorff Distance (HD) results for the task of \textbf{whole tumour segmentation}. We report the mean and standard deviation across volumes.}\label{tab1}
\resizebox{0.9\textwidth}{!}{%
\begin{tabular}{|c|cccc|cccc|}
\hline
Metrics           & \multicolumn{4}{c|}{Dice ($\uparrow$)}                                              & \multicolumn{4}{c|}{HD ($\downarrow$)}                                             \\   \hline
Pixel labels      & \multicolumn{1}{c|}{0.1\%} & \multicolumn{1}{c|}{0.5\%} & \multicolumn{1}{c|}{1\%} & 100\% & \multicolumn{1}{c|}{0.1\%} & \multicolumn{1}{c|}{0.5\%} & \multicolumn{1}{c|}{1\%} & 100\% \\ \hline
UNet              & \multicolumn{1}{c|}{$80.66_{10}$ }    & \multicolumn{1}{c|}{$86.39_{ 7.7}$}    & \multicolumn{1}{c|}{$87.34_{ 7.0}$}  &  $90.84_{ 5.6}$   & \multicolumn{1}{c|}{$9.18_{ 10}$}    & \multicolumn{1}{c|}{$6.60_{ 8.1}$}    & \multicolumn{1}{c|}{$7.37_{ 10}$}  &    $\mathbf{4.49_{ 7.2}}$ \\ \hline
SDNet             & \multicolumn{1}{c|}{$79.20_{ 11}$}    & \multicolumn{1}{c|}{$86.38_{ 7.6}$}    & \multicolumn{1}{c|}{$87.96_{ 6.6}$}  &  $\mathbf{90.96_{ 5.3}}$   & \multicolumn{1}{c|}{$11.85_{ 13}$}    & \multicolumn{1}{c|}{$7.24_{ 9.3}$ }    & \multicolumn{1}{c|}{$6.11_{ 8.4}$}  &   $4.87_{ 8.3}$   \\ \hline
vMFNet            & \multicolumn{1}{c|}{$81.30_{9.6}$}    & \multicolumn{1}{c|}{$86.14_{ 7.8}$}    & \multicolumn{1}{c|}{$87.98_{ 6.6}$ }  &  $90.62_{ 5.8}$  & \multicolumn{1}{c|}{$11.62_{ 13}$}    & \multicolumn{1}{c|}{$9.12_{ 12}$}    & \multicolumn{1}{c|}{$7.15_{ 9.6}$ }  &  $5.20_{ 8.2}$   \\ \hline
vMFBrain w/o weak & \multicolumn{1}{c|}{$79.70_{10}$}    & \multicolumn{1}{c|}{$84.92_{ 8.1}$}    & \multicolumn{1}{c|}{$87.26_{ 6.7}$}  &   $90.67_{ 5.8}$   & \multicolumn{1}{c|}{$13.89_{ 14}$}    & \multicolumn{1}{c|}{$9.80_{ 13}$}    & \multicolumn{1}{c|}{$7.18_{ 9.4}$}  &  $4.93_{ 7.3}$    \\ \hline
vMFBrain          & \multicolumn{1}{c|}{$\mathbf{85.64_{7.8}}$}    & \multicolumn{1}{c|}{$\mathbf{88.64_{ 6.8}}$}    & \multicolumn{1}{c|}{$\mathbf{89.04_{ 6.7}}$}  &   $90.58_{ 5.6}$  & \multicolumn{1}{c|}{$\mathbf{7.75_{ 7.8}}$}    & \multicolumn{1}{c|}{$\mathbf{6.18_{ 8.6}}$}    & \multicolumn{1}{c|}{$\mathbf{6.14_{ 8.4}}$}  &  $4.60_{ 6.5}$   \\ \hline
\end{tabular}
}
\end{table}

\begin{table}[t]
\centering
\caption{Dice (\%) and Hausdorff Distance (HD) results for the task of \textbf{tumour sub-region segmentation}.  We report the mean and standard deviation across volumes.} \label{tab2}
\resizebox{0.9\textwidth}{!}{%
\begin{tabular}{|c|cc|cc|cc|}
\hline
\multirow{2}{*}{0.1\% \ pixel labelled data}    & \multicolumn{2}{c|}{ED}        & \multicolumn{2}{c|}{ET}        & \multicolumn{2}{c|}{NE}       \\ \cline{2-7} 
                     & \multicolumn{1}{c|}{Dice ($\uparrow$)} & HD ($\downarrow$) & \multicolumn{1}{c|}{Dice ($\uparrow$)} & HD ($\downarrow$) & \multicolumn{1}{c|}{Dice ($\uparrow$)} & HD ($\downarrow$)  \\ \hline
UNet                 & \multicolumn{1}{c|}{$71.47_{12}$}     &  $9.60_{11}$   & \multicolumn{1}{c|}{$83.74_{8.4}$}     &  $\mathbf{5.19_{5.7}}$  & \multicolumn{1}{c|}{$79.42_{10}$}     &  $10.24_{7.9}$  \\ \hline
SDNet                & \multicolumn{1}{c|}{$75.87_{11}$}     &  $10.17_{11}$  & \multicolumn{1}{c|}{$82.45_{8.8}$}     &  $7.74_{12}$  & \multicolumn{1}{c|}{$80.70_{9.7}$}     &  $9.89_{11}$   \\ \hline
vMFNet               & \multicolumn{1}{c|}{$71.11_{12}$}     &  $10.06_{9.8}$  & \multicolumn{1}{c|}{$80.92_{9.6}$}     &  $7.99_{11}$  & \multicolumn{1}{c|}{$78.37_{11}$}     &  $12.97_{11}$  \\ \hline
vMFBrain w/o weak    & \multicolumn{1}{c|}{$70.65_{13}$}     &  $15.65_{15}$  & \multicolumn{1}{c|}{$79.36_{12}$}     &  $13.13_{17}$  & \multicolumn{1}{c|}{$79.33_{9.8}$}     &  $9.50_{9.3}$  \\ \hline
vMFBrain w/ whole tumour weak & \multicolumn{1}{c|}{$75.02_{11}$}     &  $11.56_{12}$  & \multicolumn{1}{c|}{$84.59_{8.5}$}     &  $8.17_{12}$  & \multicolumn{1}{c|}{$79.48_{9.6}$}     &  $10.02_{9.1}$  \\ \hline
vMFBrain w/ tumour sub-region weak & \multicolumn{1}{c|}{$\mathbf{78.43_{9.8}}$}     &  $\mathbf{9.14_{8.8}}$  & \multicolumn{1}{c|}{$\mathbf{85.77_{8.2}}$}     &  $5.90_{7.7}$  & \multicolumn{1}{c|}{$\mathbf{81.31_{9.0}}$}     &   $\mathbf{8.08_{7.0}}$ \\ 
\hline
\end{tabular}
}
\end{table}

\subsection{Results}

We compare model performance quantitatively using volume-wise Dice (\%) and Hausdorff Distance (95\%) (HD) \cite{dubuisson1994modified} as the evaluation metrics, and qualitatively using the interpretability and compositionality of representations. In Table~\ref{tab1} and Table~\ref{tab2}, for semi-supervised and weakly supervised approaches, the training data contains all unlabelled or weakly labelled data alongside different percentages of fully labelled data. UNet is trained with different percentages of labelled data only. Bold numbers indicate the best performance. Arrows ($\uparrow, \downarrow$) indicate the direction of metric improvement.

% \subsubsection{Brain tumour segmentation with weak labels}
\vspace{5pt} \noindent \textbf{Brain tumour segmentation with weak labels:}
Overall, as reported in Table~\ref{tab1}, the proposed vMFBrain model achieves best performance for most of the cases, particularly when very few annotations are available, i.e. the 0.1\% case. When dropping the weak supervision (vMFBrain w/o weak), we observe reduced performance, which confirms the effectiveness of weak supervision. We also observe that the reconstruction of the original image (in vMFNet) does not help. It is possible that reconstruction of the tumour does not help here because the tumour has inconsistent appearance and location between different scans. With more annotated data, all models gradually achieve better performance, as expected. Notably, with only 1\% labelled data vMFBrain achieves comparable performance (89.04 on Dice and 6.14 on HD) to the fully supervised UNet trained with all labelled data (90.84 on Dice and 4.49 on HD).

\begin{figure}[!t]
\includegraphics[width=0.9\textwidth]{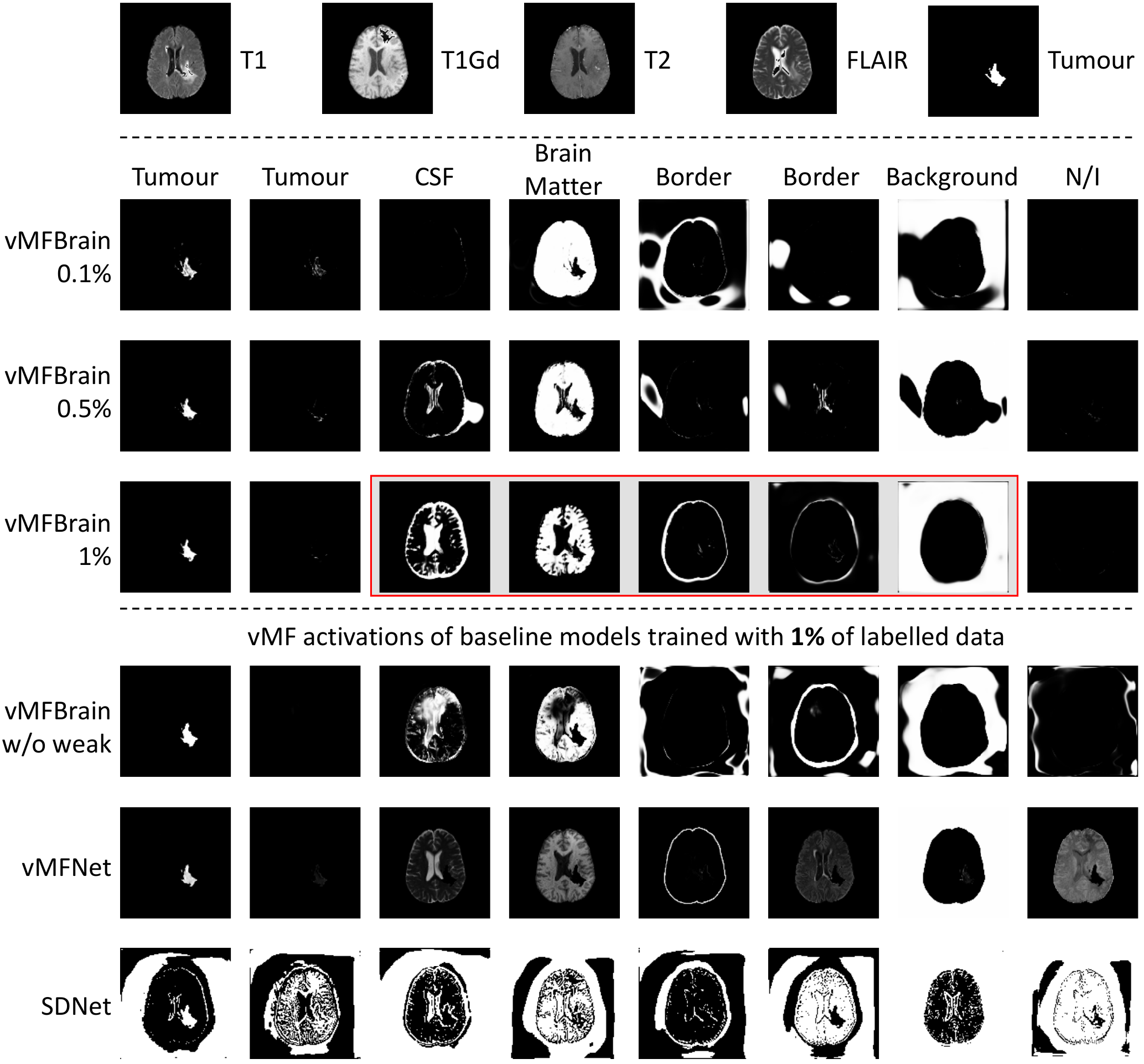}
\centering
\caption{Visualisation of vMF compositional representations (whole tumour supervision). We show the 4 input image modalities, the ground truth tumour segmentation mask, and all 8 vMF channels for the vMFBrain and baseline models trained with different percentages of labelled data. In the red boxes, the other interpretable vMF activations (excluding the tumour kernels) are highlighted. The vMF channels are ordered manually. For the vMFBrain channels, we label with a clinician's visual interpretation of which image features activated each kernel. N/I denotes non-interpretable.} \label{fig::comp}
\end{figure}

% \subsubsection{Tumour sub-region segmentation}
\vspace{5pt} \noindent \textbf{Tumour sub-region segmentation:}
We also report the results of tumour sub-region segmentation task in Table~\ref{tab2}. For this task, we perform experiments using different weak labels: a) the weak label indicating the presence of the whole tumour i.e. vMFBrain w/ whole weak and b) the weak label indicating the presence of the tumour sub-regions i.e. vMFBrain w/ sub weak. It can be seen that our proposed vMFBrain performed best with both types of weak labels. Predictably, the best performance occurs when more task-specific weak labels (i.e. weak supervision on the tumour sub-regions) are provided.

\begin{figure}[t!]
\includegraphics[width=0.9\textwidth]{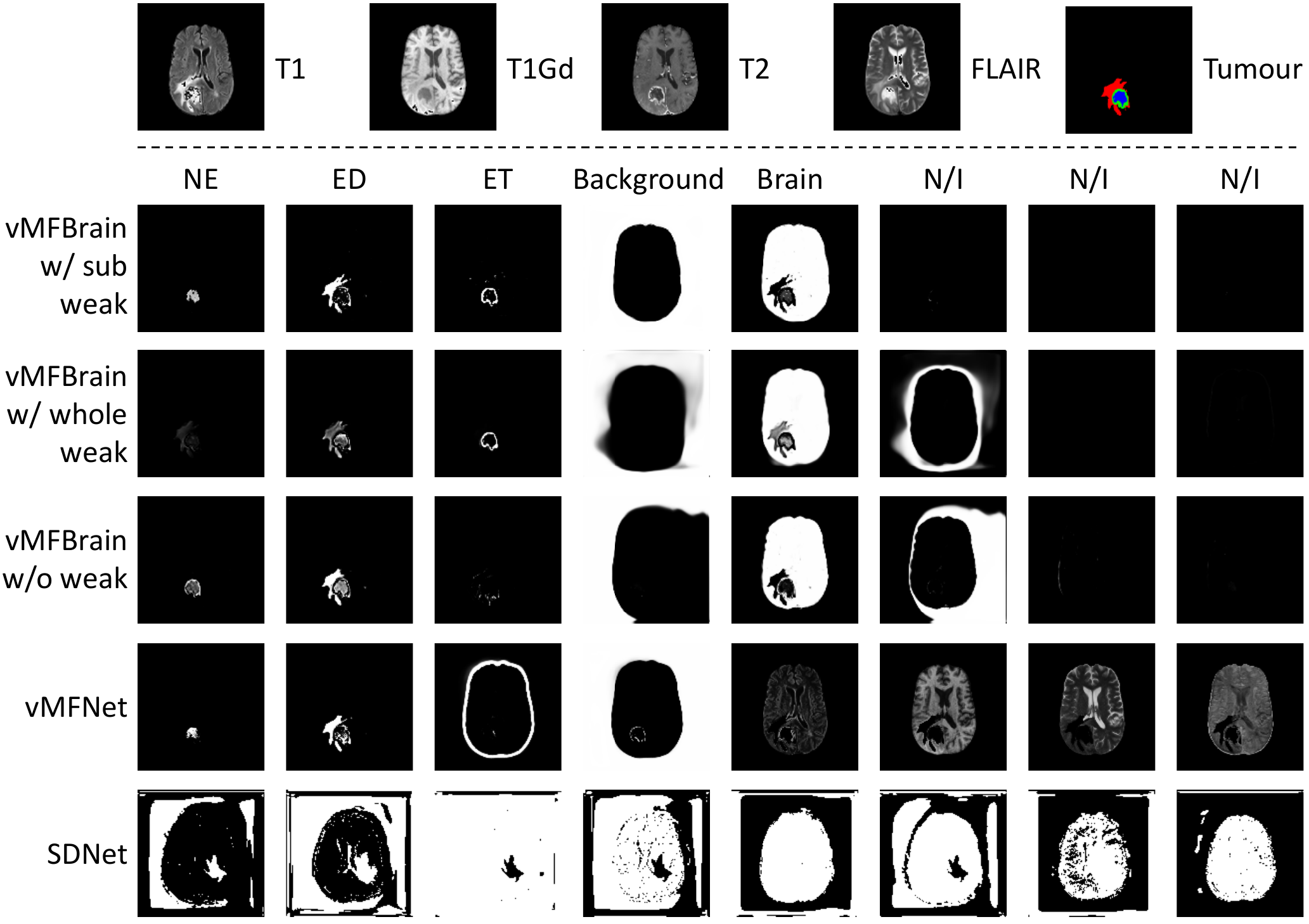}
\centering
\caption{Visualisation of vMF compositional representations (tumour sub-region supervision). We show the 4 input image modalities, the corresponding ground truth tumour sub-region segmentation mask and all 8 channels of the representations for the models trained with 1\% of labelled data. The channels are ordered manually. For the vMFBrain channels, we label with a clinician's visual interpretation of which image features activated each kernel. N/I denotes non-interpretable. } \label{fig::subcomp}
\end{figure}

% \subsubsection{Interpretability of compositional representations}
\vspace{5pt} \noindent \textbf{Interpretability of compositional representations:}
\label{sec::explain}
We are particularly interested in the compositionality of the representations when pixel labels are not sufficient. In Fig.~\ref{fig::comp}, we show the kernel activations. Note that the channels are ordered manually. For different runs, the learning is emergent such that kernels randomly learn to represent different components. Clearly, one of the kernels corresponds to the tumour in all cases. Using this kernel, we can detect and locate the tumours. For vMFBrain, training with more labelled data improves the compositionality of the kernels and the activations i.e. different kernels correspond to different anatomical or pathological structures, which are labelled by a clinician performing visual inspection of which image features activated each channel. The most interpretable and compositional representation is vMFBrain trained with 1\% labelled data. As highlighted in the red boxes, the kernels relate to CSF, brain matter, and the border of the brain even without any information about these structures given during training. Qualitatively, vMFBrain decomposes this information better into each kernel i.e. learns better compositional representations compared to other baseline models. Notably, weak supervision improves compositionality. We also show in Fig.~\ref{fig::subcomp} the representations for sub-region segmentation. Overall, we observe that with the more task-specific weak labels, the kernels learn to be more aligned with the sub-region segmentation task, where less information on other clinically relevant features is learnt. 

%\subsection{Discussion}

% \subsection{Ablation study}
% \noindent \textbf{Effect of the number of kernels:} We perform an ablation study on the number of kernels. We conduct experiments on the vMFBrain model with 6, 8 and 10 kernels for the 0.1\% labelled data case. The results are xx.xx, xx.xx and xx.xx on Dice and xx.xx, xx.xx and xx.xx on HD. 

% \noindent \textbf{Weak supervision on the latent space:} Possibly, we can perform the weak supervision on the latent representation space. We conduct the experiments by using the vMF activations as the input of the weak supervision classifier to predict the label $c$, termed vMFBrain-latent ($H=W=128$). For the 0.1\%~labelled data case, the segmentation performance  xx.xx on Dice and xx.xx on HD are much worse than the performance of vMFBrain. This supports our discussion in Section \ref{sec:method} that performing weak supervision on the representation space hurts the segmentation performance as the prediction of the presence or absence of the tumour does not require information that is essential for the tumour segmentation.

\section{Conclusion}

%It is possible to apply weak supervision on the latent space i.e. using the vMF activations as the input for the weak supervision task. However, this possibly will not help in improving the segmentation performance. For weakly-labelled data (no segmentation masks provided), the segmentation module will not be updated during training. In addition, the weak supervision task does not require the latent space to capture all the information but as little as much information for predicting the presence or absence of the tumour, which possibly hurts the segmentation performance.

In this paper, we have presented vMFBrain, a compositional representation learning framework. In particular, we constructed weak labels indicating the presence or absence of the brain tumour and tumour sub-regions in the image. Training with weak labels, better compositional representations can be learnt that produce better brain tumour segmentation performance when the availability of pixel-level annotations is limited. Additionally, our experiments show the interpretability of the compositional representations, where each kernel corresponds to specific anatomical or pathological structures. Importantly, according to our experiments and the results reported in previous studies \cite{liu2022vmfnet, liu2023compositionally}, the vMF-based compositional representation learning framework is robust and applicable to different medical datasets and tasks. In future work, we might consider transferring vMFBrain to 3D in order to process wider spatial context for each structure.

\vspace{5pt} \noindent \textbf{Acknowledgements}
S.A. Tsaftaris acknowledges the support of Canon Medical and the Royal Academy of Engineering and the Research Chairs and Senior Research Fellowships scheme (grant RCSRF1819\textbackslash8\textbackslash25). Many thanks to Patrick Schrempf and Joseph Boyle for their helpful review comments.

\bibliographystyle{splncs04}
\bibliography{references}

\end{document}